\documentclass[10pt,twocolumn,letterpaper]{article}

\usepackage{btas}
\usepackage{times}
\usepackage{epsfig}
\usepackage{graphicx}
\usepackage{amsmath}
\usepackage{amssymb}
\usepackage[norule,symbol,perpage]{footmisc}

\usepackage[pagebackref=true,breaklinks=true,letterpaper=true,colorlinks,bookmarks=false]{hyperref}

\btasfinalcopy % *** Uncomment this line for the final submission
\newcommand{\tabincell}[2]{\begin{tabular}{@{}#1@{}}#2\end{tabular}}

\ifbtasfinal\fi

\makeatletter
\def\ps@IEEEtitlepagestyle{
\def\@oddfoot{\mycopyrightnotice}
\def\@evenfoot{}
}
\def\mycopyrightnotice{
{\hfill \footnotesize 978-1-7281-1522-1/19/\$31.00 \copyright 2019 IEEE \hfill}
}
\makeatother

\begin{document}

%%%%%%%%% TITLE
\title{Cosmetic-Aware Makeup Cleanser}
%(for Makeup-Invariant Face Verification)
\author{ Yi Li, Huaibo Huang, Junchi Yu, Ran He, Tieniu Tan\\
CRIPAC, National Laboratory of Pattern Recognition, CASIA \\
%Center for Research on Intelligent Perception and Computing, CASIA \\
Center for Excellence in Brain Science and Intelligence Technology, CAS \\
University of Chinese Academy of Sciences \\
{\tt\small \{yi.li,huaibo.huang\}@cripac.ia.ac.cn, rhe@nlpr.ia.ac.cn}
%\and
}

\maketitle
\thispagestyle{empty}

%%%%%%%%% ABSTRACT
\begin{abstract}
   Face verification aims at determining whether a pair of face images belongs to the same identity.
   Recent studies have revealed the negative impact of facial makeup on the verification performance.
   With the rapid development of deep generative models, this paper proposes a semantic-aware makeup cleanser (SAMC) to remove facial makeup under different poses and expressions and achieve verification via generation.
   The intuition lies in the fact that makeup is a combined effect of multiple cosmetics and tailored treatments should be imposed on different cosmetic regions.
   To this end, we present both unsupervised and supervised semantic-aware learning strategies in SAMC.
   At image level, an unsupervised attention module is jointly learned with the generator to locate cosmetic regions and estimate the degree.
   At feature level, we resort to the effort of face parsing merely in training phase and design a localized texture loss to serve complements and pursue superior synthetic quality.
   %The simple yet effective method is implemented by a new generative network which is trained in an end-to-end way.
   The experimental results on four makeup-related datasets verify that SAMC not only produces appealing de-makeup outputs at a resolution of $256 \times 256$, but also facilitates makeup-invariant face verification through image generation.
\end{abstract}

\let\thefootnote\relax\footnotetext{\mycopyrightnotice}

%%%%%%%%% BODY TEXT
\section{Introduction}
As one of the most representative regions of a human being, face acts as a vital role in biometrics.
The appearance of a face is often deemed as a crucial feature for identifying individuals.
Most face verification methods determine whether a pair of face images refers to the same person by comparing certain facial features from the given images.
Following the remarkable progress of face verification research \cite{sun2013hybrid,taigman2014deepface,sun2014deep,jing2016multi,zhang2016multi,he2017learning,wu2018light}, the related technology brings great convenience to our lives ranging from social media to security services.
Nevertheless, it is worth noticing that most of these application scenarios are concerned in information security and the importance of algorithm reliability is self-evident.
Facing various special situations, there is still a long way to solve the problem thoroughly.

\begin{figure}[t]
  \centering
  \includegraphics[width=0.85\linewidth]{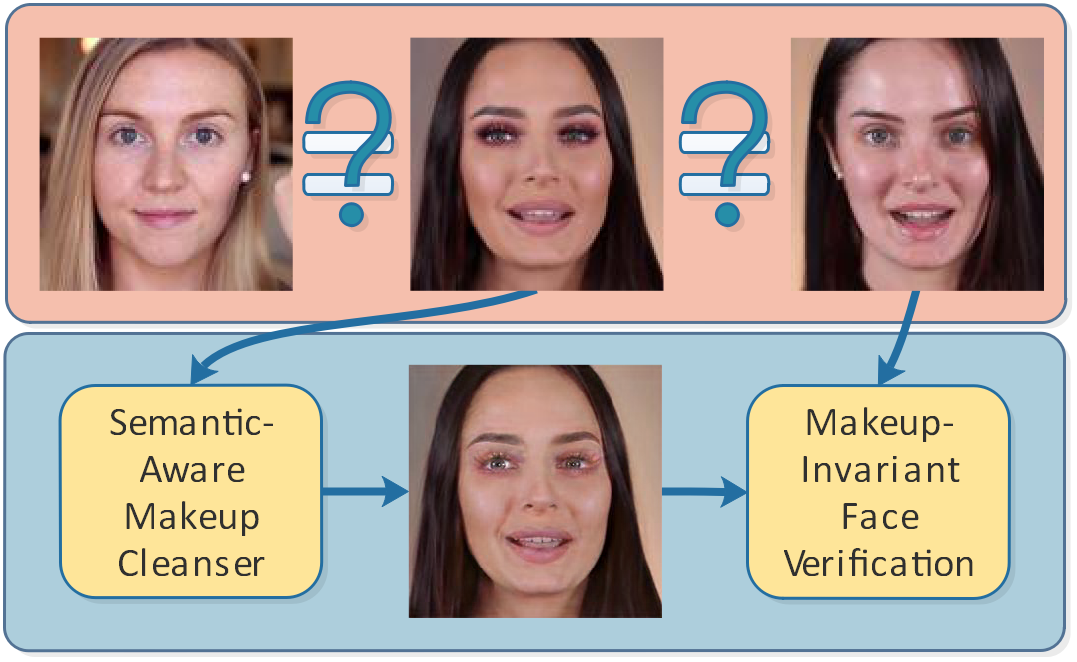}\\
  \caption{Conceptual illustration of verification via generation. 
  %Makeup is applied intentionally to change one's appearance, which poses impacts on face verification. 
  The makeup problem is quite challenging since it often jointly occurrs with pose and expression variations. A semantic-aware makeup cleanser is proposed to remove makeup while keep other factors unchanged
  %, making the generated non-makeup image benefit makeup-invariant face verification
  .}
  \label{idea}
\end{figure}

Although face is an inherent surface area of a person, its appearance on images may alter due to many factors, e.g., view angles, expressions and makeup changes.
Among these factors, the makeup change has caught more attention than ever because of the striking development of facial cosmetics and virtual face beautification.
Makeup is usually used to conceal flaws on the face, enhance attractiveness or alter appearance.
Not only females, there are more and more males that agree makeup being a daily necessity and even a courtesy on certain occasions.
Comparing to view angle and expression changes that are mostly out of uncooperative behaviors, makeup is a special case that may be inevitable during photographing.
However, the appearance changes caused by makeup will also decrease the verification performance, just like other factors.
This paper studies to remove facial makeup even under cases with pose and expression variations.

A typical facial makeup style can be divided into three main parts according to their locations: eye shadows, lipstick and foundation \cite{chang2018pairedcyclegan,li2018beautygan}.
The comprehensive effects of various cosmetics can lead to significant changes on facial appearance.
Since there are theoretically innumerable kinds of makeup styles, the nondeterminacy and variability make it quite difficult to match the images before and after makeup.
Besides, makeup is different from plastic surgery for it concerns non-permanent changes and is easy to be cleansed.
In contrast to skin care products, cosmetics are regarded as a harm to skin health and thus most people only wear makeup when meeting others.
The convenient facial changes induced by makeup are reported in \cite{dantcheva2012can} to have posed a severe impact on face verification systems (which often depend much on capturing various cues from facial appearance).
In this paper, we study the effect of makeup and contrive to alleviate its impacts on face verification performance in a generative way.

To address makeup-invariant face verification, much efforts have been made in the past decade and we sum them up in two main streams.
Early methods concentrate on exacting discriminative features directly from input images.
By maximizing correlation between images of the same subjects, these methods map images into a latent space for better classification performance.
The mapping function can be designed manually \cite{hu2013makeup,guo2014face,chen2016ensemble} or learned by a deep network \cite{sun2017weakly}.
Another stream resorts to the success of deep generative models and uses the makeup removal outputs for verification, represented by \cite{li2017anti,li2019learning}.
Given a face image with makeup, these methods first generate a non-makeup and identity-preserving image based on the input.
And then the generated images along with real non-makeup images are utilized for verification.
Since this stream makes changes at image level, it achieves the advantage of adapting existing verification methods for makeup problems with no requirement of retraining.

However, \cite{li2017anti,li2019learning} formulate makeup removal as a one-to-one image translation problem, which in fact lacks rationality.
During training, there is a pair of makeup and non-makeup images of a subject, the networks in \cite{li2017anti,li2019learning} force the output to be like the non-makeup ground truth instead of just removing facial makeup.
We show a sample pair in Figure~\ref{idea} and obvious data misalignment can be observed.
Apart from makeup, many other factors are also different in two images, including the hair style and the background.
In this paper, we argue that makeup removal is naturally a task with unpaired data for it is illogical to obtain a pair of images with and without makeup simultaneously.
When removing the makeup, other factors of an output are expected to retain the same with its input, based on the premise of realistic visual quality.

To this end, we propose a Semantic-Aware Makeup Cleanser (SAMC) to facilitate makeup-invariant face verification via generation.
As mentioned above, facial makeup is the results of applying multiple cosmetics and different regions can appear different effects.
Thus we raise the idea of removing the cosmetics with tailored schemes.
Concretely, we adopt two semantic-aware learning strategies in SAMC, in both unsupervised and supervised manners.
For an input image with makeup, we first utilize an attention module to locate the cosmetics and estimate their degrees.
The attention module is jointly learned with the generator in an unsupervised way, since there is no access to applicable attention maps serving as supervision.
The aim of this process is to get explicit knowledge of makeup automatically as well as adaptively.
Although the attention module suggests where to stress, its confidence and accuracy are not as satisfying as desired, especially at the beginning of the training phase.
To address this issue, we propose a semantic-aware texture loss to set a constraint on the synthesized texture of different cosmetic regions.
In addition, the misalignment problem is overcome by a feat of face image warping.
By warping the non-makeup ground truth according to its face keypoints, we obtain pixel-wise alignment data, benefiting precise supervision and favorable generation.

%We conduct experiments on four datasets that are commonly used for makeup-invariant face verification.
%The results verify that SAMC profits verification performance by generating high-quality non-makeup images.
%To sum up, the major contributions of this paper are as follows.
%\begin{itemize}
%  \item A  Semantic-Aware Makeup Cleanser (SAMC) is proposed to remove facial makeup with two elaborate schemes, in both unsupervised and supervised manners. SAMC is implemented by a newly designed generative network trained in an end-to-end way.
%  \item At image level, an attention module is learned in an unsupervised strategy to suggest where and how the cosmetics lie. At feature level, we design a semantic-aware texture loss to serve supervision with better confidence and accuracy.
%  \item Extensive experiments (involving various poses and expressions) demonstrate that SAMC not only produces visually appealing makeup removal outputs (at a resolution of $256 \times 256$) but also boosts face verification via generation.
%\end{itemize}

%\begin{table}
%\begin{center}
%\begin{tabular}{|l|c|}
%\hline
%Method & Frobnability \\
%\hline\hline
%Theirs & Frumpy \\
%Yours & Frobbly \\
%Ours & Makes one's heart Frob\\
%\hline
%\end{tabular}
%\end{center}
%\caption{Results.   Ours is better.}
%\end{table}

\begin{figure*}[t]
  \centering
  \includegraphics[width=0.9\linewidth]{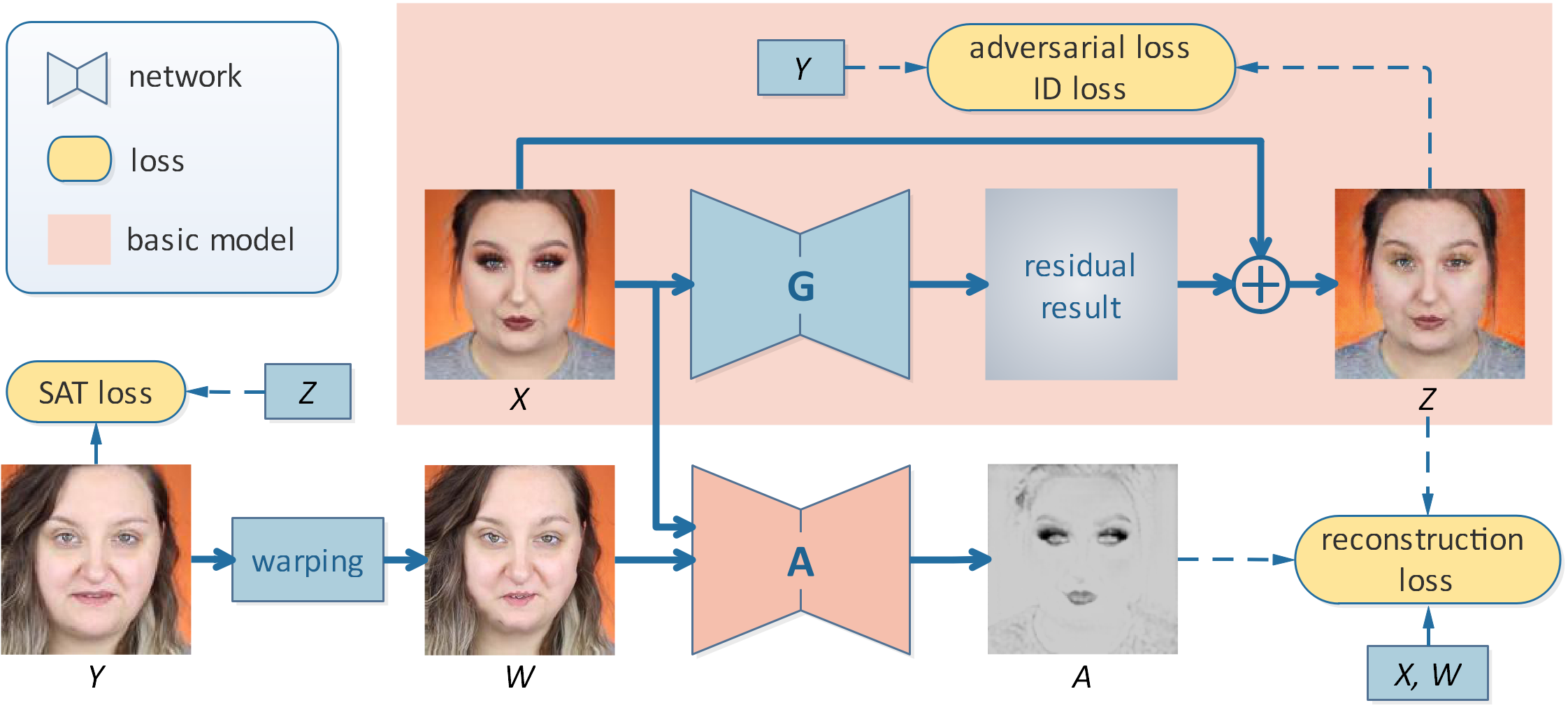}\\
  \caption{The network diagram of the Semantic-Aware Makeup Cleanser (SAMC). $X$ and $Y$ are a pair of images with and without makeup belonging to the same subject. Obvious misalignment can be observed. Instead of directly using $Y$ as the ground truth, we first warp $Y$ according to $X$ and obtain $W$ for better supervision. Then an unsupervised attention module is learned to locate the makeup and estimate the degree. The attention map $A$ indicates where and how much should be changed. Finally, we propose a SAT loss to constrain the local texture of the generated result $Z$. The generator receives four types of losses in the full structure, while the attention module is merely learned by the reconstruction loss.}
  \label{SAMC}
\end{figure*}

\section{Approach}

\subsection{Motivation and Overview}
In reality, makeup images are generally acquired in unrestricted conditions, i.e., in the wild.
Besides makeup, there exist complex variations including pose, expression, illumination.
Although these factor may also impact verification performance and should be adjusted, the proposed SAMC focuses on makeup removal but leaves out other factors for better user experience.
When a customer is using a makeup removal system, the changes are expected to be limited to makeup.
Therefore, given an image with makeup as the input, our method aims at removing the makeup in a generative manner while retaining other information.
The output is expected to achieve appealing visualization as well as suffice face verification.
%The process can be considered as an unsupervised image-to-image translation problem, but with distinct characteristics.
%Pix2pix \cite{isola2016image} and CycleGAN \cite{Zhu2017Unpaired} are two representative universal image-to-image translation methods.
%In theory, they can also be applied to makeup removal.
%However, the makeup problem lacks exactly aligned data which is required to train Pix2pix.
%Although CycleGAN fulfills image translation with unpaired data, it is inadequate for makeup removal.
%The major reason lies in that CycleGAN considers the input image as a whole and thus is inherently unable to grasp the subtle texture changes before and after makeup.
Different from image style transfer, the effect of cosmetics merely involves certain face areas instead of the global image.
Moreover, despite the inaccessibility of exactly aligned data, we can acquire image pairs of a subject, one with makeup and the other without.
With proper processes and strategies, these image pairs can provide superior supervision.

To fully utilize the image pairs without exact alignment, we propose a semantic-aware makeup cleanser to remove facial cosmetics with tailored strategies.
The simple yet effective method is implemented by a network that mainly contains a generator $G$ and an attention module $A$.
We employ image pairs $\{X_{1}, Y_{1}\}, \{X_{2}, Y_{2}\} \cdots$ to train the network and assume that $X \in \mathcal{X}$ and $Y \in \mathcal{Y}$ represent images with and without makeup.
It is noting that $\{X_{i}, Y_{i}\}$ refer to the same identity but may differ in expression, pose, background, etc.
To obtain pixel-wise supervision, we warp the non-makeup ground truth $Y_{i}$ according to $X_{i}$, yielding the warped non-makeup ground truth $W_{i}$.
The warping process consists two steps: 1) detecting 68 facial keypoints by \cite{bulat2017far}, and 2) non-linear transformation in \cite{ruprecht1995image}.
In the following, we will elaborate the network and loss functions in details.

\subsection{Basic Model}
We first describe the basic model of SAMC which is typically a conditional generative adversarial network \cite{goodfellow2014generative,isola2016image} with an identity constraint, similar as \cite{li2017anti}.
The diagram of the basic model lies in the top right corner of Figure \ref{SAMC}.
Taking $X \in \mathbb{R}^{w \times h \times 3}$ as the input, the basic model produces the makeup removal result $Z\in \mathbb{R}^{w \times h \times 3}$ though the generator $G$.
Different from \cite{li2017anti} that forces $Z$ to resemble $Y$, we expect the changes to be limited in cosmetic areas while other regions are kept the same as the input.
Therefore, the U-net structure \cite{ronneberger2015u,isola2016image} is adopted in the generator $G$ for its skip connections help to maintain abundant context information during the forward pass.
Instead of mapping to the output $Z$ directly, the network learns a residual result as a bridge, inspired by the success of ResNet \cite{he2016deep}.
The final output $Z$ is obtained by adding the residual result to the input $X$.

The generator $G$ receives two types of losses to update parameters in the basic model, i.e., an adversarial loss and an identity loss.
The vanilla GAN \cite{goodfellow2014generative} uses the idea of a two-player game tactfully to achieve the most promising synthetic quality.
The two players are a generator and a discriminator that compete with each other.
The generator aims at producing samples to full the discriminator, while the discriminator endevours to tell real and fake data apart.
To refrain from artifacts and blurs in the output, we train a discriminator to serve the adversarial loss which can be formulated as
\begin{equation}\label{adv}
  L_{adv} = \mathbb{E}_Y [log D(Y)] + \mathbb{E}_{X,Y} [log(1- D(Z))].
\end{equation}

In addition to removing makeup, we also expect that the generated images can maintain the identity of the original image, contributing to improve the verification performance across makeup status.
Different from visual quality, verification performance is calculated by comparing image features extracted by tailored schemes.
For face verification, the key issue is to generate images with qualified features that indicates identity.
Thus we also use the identity loss to keep the identity information consistent.
The identity loss is used as a classical constraint in a wide range of applications, e.g., super-resolution \cite{johnson2016perceptual}, face frontalization \cite{cao2018learning}, age estimation \cite{li2018global}, and makeup removal \cite{li2017anti}.
Similar to \cite{li2017anti}, we employ one of the leading face verification networks, i.e. Light CNN \cite{wu2018light}, to obtain the identity information of an face image.
The ID loss is calculated by
\begin{equation}\label{ID}
  L_{ID} =  E_{Y,Z} {\left\| {F(Z) - F(Y)} \right\|_2}
\end{equation}
where $F(\cdot)$ represents the pre-trained feature extractor.
Noting that the parameters in $F(\cdot)$ stay constant during training.

\subsection{Attention Module}
%Makeup is the effect of multiple cosmetics applied to a human face.
%It is widely used to improve attraction or disguise oneself.
By comparing images before and after makeup of a certain subject, we observe that the facial changes caused by makeup only concern several typical regions, like eye shadows and lips.
Existing makeup removal methods like \cite{li2017anti} treat the problem as a one-to-one image translation task and force the output to resemble the non-makeup image in the dataset.
This behavior violates the observation above and thus is not capable of generating satisfying results.
Instead, when removing makeup, we should concentrate on these cosmetic regions and leave other unrelated image areas out.
In this paper, an attention module is developed to enhance the basic model by distinguishing where and how much are the cosmetics in a pixel-wise way.

On the other hand, the attention map can also contribute to ignore the severe distortion in the warped non-makeup image $W$ as shown in Figure~\ref{SAMC}.
The aim of the warping process is to alleviate data misalignment and provide better supervision.
However, the warping is based on matching the facial keypoints on two images and the distortion is somewhat inevitable after the non-linear transformation.
If provided with distortion-polluted supervision, the generator may be misguided and produce results with fallacious artifacts and distortion.
To mitigate this risk, we adopt the attention map to adaptively decide where to believe in the warped image.
We present a learned attention map $A \in \mathbb{R}^{w \times h}$ in Figure~\ref{SAMC}.
%The attention map shares the same height and width with the input but has just one channel, i.e., $A \in \mathbb{R}^{w \times h}$.
The element in $A$ is normalized between $0$ and $1$.
In Figure~\ref{SAMC}, we adjust the colour of $A$ for better visualization.
It can be concluded that the makeup focuses on eye shadows and lips, in accord with common sense.
In general, there are dark and light pixels in $A$.
The dark ones indicate that the makeup removal result at this pixel should be like $W$, and the light pixel should be like $X$.
In this way, the distortion pixels are successfully neglected owing to their weights close to 0.
We formulate the intuition as the reconstruction loss, which is calculated by
\begin{equation}\label{rec}
  L_{rec} = | A \otimes W - A \otimes Z | + | (1-A) \otimes X - (1-A) \otimes Z |
\end{equation}
where $\otimes$ represents element-wise multiplication between matrices.

Nevertheless, there is another problem that the attention map is easy to converge to all $0$ or all $1$.
When the attention map is all $0$, it means that $Z$ is driven to be the reconstruction of $A$.
If the attention map is all $1$, $Z$ is induced to imitate $W$.
Neither case is expected to occur.
To avert these cases, \cite{Mejjati2018Unsupervised} introduces a threshold to control the learning effect of the attention map.
However, it is difficult to choose an appropriate value for the threshold.
Different from it, we employ a simple regularization to address the problem, which is implemented by the additional L1 norm constraint between the two terms in Equation~\ref{rec}.
This regularization prevents the two terms from being too large or too small, and thus restricts the value in the attention map consequentially.
A balanced weight is introduced to control the contribution of the regularization and we empirically set it as $0.2$.

\begin{figure}
  \centering
  \includegraphics[width=0.8\linewidth]{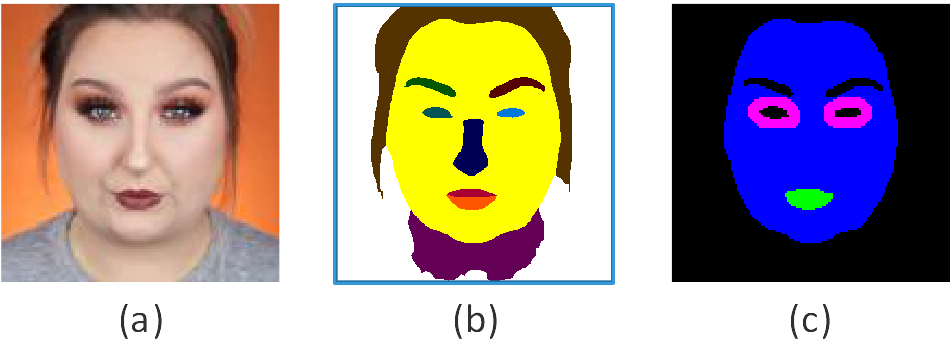}\\
  \caption{A face parsing sample. Taking (a) as the input, \cite{wei2017learning} produces the face parsing result in (b). We make further modification and obtain (c), for calculating SAT loss.}\label{parsing}
\end{figure}

\subsection{Semantic-Aware Texture Loss}
The attention map is learned along with the generator to indicate makeup at image level.
Considering that it is learned in an unsupervised manner, the confidence and accuracy cannot be ensured, especially at the beginning of the training process.
Hence, we explore other supervision to pursue better generation quality and stable network training.
To this end, a Semantic-Aware Texture (SAT) loss is proposed to make the synthesized texture of different cosmetic regions realistic.
As has been analyzed, makeup is substantially a combination of cosmetics applied to multiple facial regions.
A typical makeup effect can be divided into foundation, eye shadows and lipsticks. Based on these, we resort to the progress of face parsing \cite{wei2017learning} and further adapt the parsing results to obtain different cosmetic regions.
Figure~\ref{parsing} presents a set of parsing results. There are three colours in Figure~\ref{parsing}(c), each standing for a cosmetic region.

The aim of the SAT loss is to resemble the local texture of $Z$ to that of $Y$.
After acquiring the cosmetic region label map of $X$ and $Y$, the mean $\mu$ and standard deviation $\sigma$ of each feature region is calculated accordingly.
Noting that 1) we assume that $Z$ shares the label map with $X$ for their appearance merely differs in makeup effects, and 2) the label map has been resized according to the corresponding feature map.
As for the texture extractor, we continue to adopt $F(\cdot)$ but only use the output of the second convolutional layer. Finally, the SAT loss is defined as
\begin{equation}\label{SAT}
  L_{sat} = \sum_i \| \mu_{i}^{Z} - \mu_{i}^{Y} \|_2 +  \| \sigma_{i}^{Z} - \sigma_{i}^{Y} \|_2
\end{equation}
where $\mu_{i}^{*}$ and $\sigma_{i}^{*}$ represent the mean and standard deviation of $*$ with $i = \{1,2,3\}$ indicating the three cosmetic regions.

In total, the generator updates its parameters based on the elaborated four losses and the full objective is
\begin{equation}\label{obj}
  L_G \triangleq L_{adv} +  L_{ID} +  L_{rec} + L_{sat}.
\end{equation}

\section{Experiments}
\begin{figure}
  \centering
  \includegraphics[width=\linewidth]{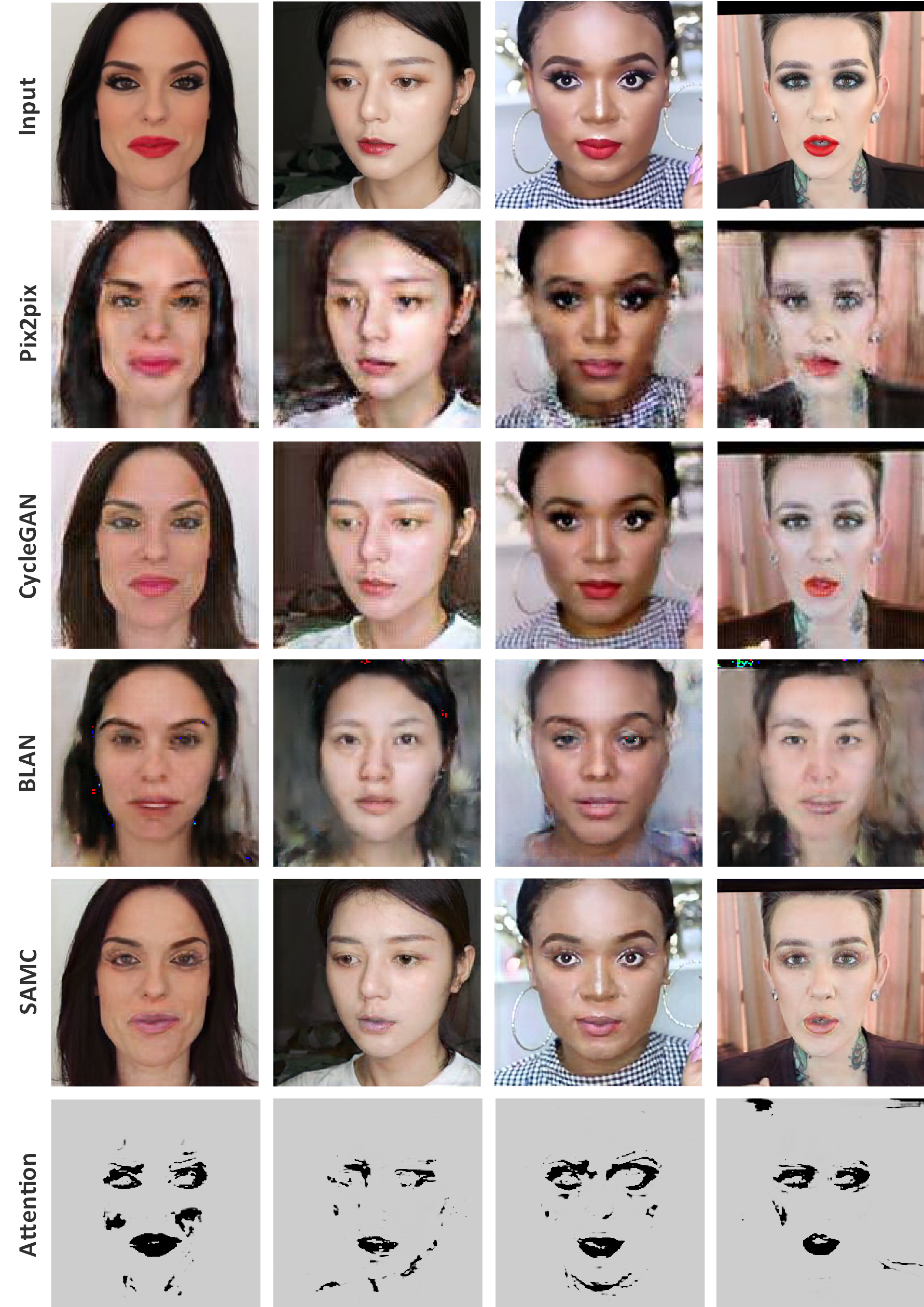}\\
  \caption{Visualization of makeup removal and the attention map obtained by SAMC.
  %Makeup removal in the wild is challenging due to various poses and expressions.
  The best results are generated by our method on the right. SAMC sucessfully cleanses the facial cosmetics, especially in regions like eye shadows and lips. For better visual effects, the attention map is processed by reducing brightness and the black pixels locate the cosmetics successfully.
  }\label{de-makeup}
\end{figure}
%
%We perform experiments on four datasets to verify the effectiveness of SAMC for makeup-invariant face verification.
%In addition, our network structure can handle high-resolution images, providing inspiration for further related research.
%The visualizations of synthetic non-makeup images and the comparison with existing methods are shown in Figure \ref{de-makeup}.
%Moreover, analysis experiments are carried out based on the design of the network and the use of the loss function.

\subsection{Datasets and Training Details}
We use three public datasets to build our training and test sets.
\textbf{Dataset1} is collected in \cite{guo2014face} and contains 501 identities.
For each identity, a makeup image and a non-makeup image are included.
\textbf{Dataset2} is assembled in \cite{sun2017weakly} and contains 406 images.
Similarly, each identity includes a makeup image and a non-makeup image.
\textbf{Dataset3 (FAM)} \cite{hu2013makeup} consists of 519 pairs of images.
For fair comparison with other methods, we resize all the images to the resolution of $128\times128$ in our experiments.
%
%Since the mentioned three datasets are all collected by searching for eligible images on the Internet directly, there are problems of inconsistency of identity age, imaging condition, etc., as well as blur and noise.
%Furthermore, the images are of low resolution (typically below $256 \times 256$), which appears to be inadequate for the development of image generation.
We also employ the high-quality makeup data collected by our lab to train and test the network.
We refer to it as the \textbf{Cross-Makeup Face (CMF)} dataset.
%Inspired by \cite{chang2018pairedcyclegan}, we first search for makeup tutorial videos on the Internet.
%For each video, two frames with a distinct and non-occlusion face are extracted, one for before makeup and the other for after makeup.
%Significant differences can be observed between makeup and non-makeup images.
There are 2600+ image pairs at high resolution (at least $256 \times 256$) of 1400+ identities, involving both makeup variations and identity information.

The experiments involve images with two resolutions ($128\times128$ and $256\times256$).
Our network is implemented based on pytorch.
We train SAMC on the CMF dataset and test it on all the four datasets mentioned above.
Therefore, the experiments on Dataset 1--3 are conducted in a cross-dataset setting.
There are two main subnetworks in SAMC, i.e., the generator $G$ and the attention module $A$.
In the implementation, $G$ and $A$ share the same architecture but update parameters separately.
As for the feature extractor $F(\cdot)$ in Eq. \ref{ID} and \ref{SAT}, we utilize the released model of Light CNN which is trained on MS-Celeb-1M \cite{guo2016ms}.
The batch size is set to $8$.
The model is trained using the Adam algorithm with a learning rate of $0.0001$.
We set the balanced weights of all the losses as $1$ without loss of generality.

\begin{table}[t]
\begin{center}
\caption{The verification results(\%) on the CMF dataset.}
\label{tab-cmf}
\begin{tabular}{|c|ccc|}
  \hline
  Method  &   Rank-1  & \tabincell{c}{TPR@FPR \\ =0.1\%} &  \tabincell{c}{TPR@FPR \\ =1\%}  \\
  \hline \hline
  baseline &    $91.92$  &  $63.29$  & $86.16$   \\
  Pix2pix \cite{isola2016image}&    $76.54$   &  $36.24$    &  $69.34$\\
  CycleGAN \cite{Zhu2017Unpaired}&    $88.65$   &  $57.00$    &  $82.78$\\
  BLAN \cite{li2017anti}       &   $91.03$  &  $55.62$  & $85.29$   \\
   ours &      $94.04$   &  $77.36$    &  $93.55$\\
  \hline \hline
   ours \verb|\| ID &      $90.58$   &  $63.44$    &  $86.40$\\
   ours \verb|\| SAT &      $91.54$   &  $65.96$    &  $87.50$\\
   ours \verb|\| adv &      $91.35$   &  $65.64$    &  $87.42$\\
   ours-att\_0 &      $91.15$   &  $66.27$    &  $87.42$\\
   ours-att\_1 &      $91.73$   &  $64.86$    &  $86.95$\\
  \hline
\end{tabular}
\end{center}
\end{table}

\begin{table*}[t]
\begin{center}
\caption{Rank-1 accuracy (\%) on three released makeup datasets.}
\label{tab-fam}
\begin{tabular}{|c|cccccccc|}
  \hline
  Method         &  \cite{guo2014face} & \cite{sun2017weakly} & \cite{nguyen2010cosine} &  \cite{hu2013makeup} & VGG \cite{simonyan2014very} & LightCNN \cite{wu2018light} & BLAN \cite{li2017anti} &  ours \\
  \hline \hline
  Dataset 1  & 80.5 & 82.4 &  -  &  -  & 89.4 & 92.4 & 94.8 &  96.1 \\
  Dataset 2  &  -  & 68.0 &  -  &  -  & 86.0 & 91.5 & 92.3 &  96.7 \\
  FAM        &  -  &  -  &  59.6  &  62.4  & 81.6 & 86.3 & 88.1  & 94.8 \\
  \hline
\end{tabular}
\end{center}
\end{table*}

\subsection{Visualization of Makeup Removal}
Figure \ref{de-makeup} presents some sample results of makeup removal, along with the learned attention of SAMC.
Noting that we modify the display effect of attention maps for better visualization.
We compare our results with other methods, including Pix2pix \cite{isola2016image}, CycleGAN \cite{Zhu2017Unpaired} and BLAN \cite{li2017anti}.
Pix2pix and CycleGAN are widely considered as representative methods in supervised and unsupervised image-to-image translation, respectively.
To the best of our knowledge, BLAN firstly propose to generate non-makeup images from makeup ones for makeup-invariant face verification.
We train these networks from scratch on CMF training data and the configurations are kept the same as described in their papers.

In Figure \ref{de-makeup}, it can be observed that Pix2pix and BLAN generates images with severe artifacts and distortion.
The reason lies in that these methods assume the existence of well aligned data pairs and formulate the image translation problem at pixel level.
As has been analyzed, makeup images inherently lack paired data, making the problem more difficult to tackle with.
Although CycleGAN is learned in an unsupervised manner and produces images of higher quality than its neighbours, there exist apparent cosmetic residues due to the lack of proper supervision.
As for the attention map, it demonstrates that our attention module can locate the cosmetics.
And we will discuss the contribution of the attention module in the ablation studies.
Comparing with other methods, our network achieves the most promising results.
Not only are the cosmetics successfully removed, but also other image details are well preserved.

\subsection{Makeup-Invariant Face Verification}

In this paper, we propose a makeup removal method in the aim of facilitating makeup-invariant face verification via generation.
To quantitatively evaluate the performance of our makeup cleanser, we show the verification results of SAMC and related algorithms in Table \ref{tab-cmf}.
As introduced above, we adopt the released Light CNN model as the feature extractor.
Concretely, each image goes through the Light CNN and becomes a $256$-d feature vector.
The similarity metric used in all experiments is cosine distance.
For the baseline, we use the original images with and without makeup as inputs.
For other methods, we instead use the makeup removal results and the original non-makeup images as inputs for verification.

For the CMF dataset, we observe that our approach brings significant improvements on all the three criteria.
Instead of forcing the output to resemble the ground truth non-makeup image in the dataset like BLAN, we learn to locate and remove the cosmetics while maintaining other information including pose and expression.
The accuracy improvements demonstrate that our network alleviates the side effects of makeup on face verification by generating high-quality images.
On the other hand, CycleGAN fails to preserve identity information during generation, even though it produces outputs with moderate quality.
The reason is that Pix2pix and CycleGAN are designed for general image translation and take no discrimination into account.
For fair comparison, we further conduct experiments on three public makeup datasets and the results are exhibited in Table \ref{tab-fam}.
It is worth noticing that BLAN is trained on these datasets while our SAMC is trained on CMF and tested on these without adaptation.
Thanks to the stability of our network, SAMC can still outperform other methods in cross-dataset settings.

Besides, ablation studies are conducted to evaluate the contribution of each component in SAMC. 
We remove or modify one of the components and concern the changes in the corresponding metrics to verify their importance.
To study the used loss functions to train the generator, we build three variants: 1) training without the ID loss, 2) training without the SAT loss, and 3) training without the adversarial loss.
Since the reconstruction loss involves the learned attention map, we utilize different attention schemes to analyze the effect of the attention module.
In particular, the attention map is set to all $0$ and all $1$, respectively.
The quantitative verification results are reported in Table \ref{tab-cmf} for comprehensive and fair comparison.
The ``$\verb|\|$'' and the ``w$\verb|\|$o'' represent ``without''.
As expected, the performance of removing either component will experience a drop.
By observing the accuracies in Table \ref{tab-cmf}, we can find that there is an apparent decline when removing the ID loss.
It indicates the effectiveness and importance of the ID loss in preserving discriminative information at feature level.

\section{Conclusion}
In this paper, we focus on the negative impact of facial makeup on verification and propose a semantic-aware makeup cleanser (SAMC) to remove cosmetics. Instead of considering makeup as an overall effect, we argue that makeup is the combination of various cosmetics applied to different facial regions. Therefore, a makeup cleanser network is designed with integration of two elaborate schemes. At image level, an attention module is learned along with the generator to locate the cosmetics in an unsupervised manner. Specifically, the elements in the attention map range from $0$ to $1$ with different values indicating the makeup degree. At feature level, a semantic-aware texture loss is designed to serve complements and provide supervision. Experiments are conducted on four makeup datasets. Both appealing makeup removal images and promising makeup-invariant face verification accuracies are achieved, verifying the effectiveness of SAMC.

\section{Acknowledgments}
{This work is partially funded by National Natural Science Foundation of China (Grant No. 61622310), Beijing Natural Science Foundation (Grant No. JQ18017), and Youth Innovation Promotion Association CAS (Grant No. 2015109).}

{\small
\bibliographystyle{ieee}
\bibliography{bibfile}

\begin{thebibliography}{10}\itemsep=-1pt

\bibitem{bulat2017far}
A.~Bulat and G.~Tzimiropoulos.
\newblock How far are we from solving the 2d \& 3d face alignment problem?(and
  a dataset of 230,000 3d facial landmarks).
\newblock In {\em Proceedings of the IEEE International Conference on Computer
  Vision}, pages 1021--1030, 2017.

\bibitem{cao2018learning}
J.~Cao, Y.~Hu, H.~Zhang, R.~He, and Z.~Sun.
\newblock Learning a high fidelity pose invariant model for high-resolution
  face frontalization.
\newblock In {\em Advances in Neural Information Processing Systems}, pages
  2872--2882, 2018.

\bibitem{chang2018pairedcyclegan}
H.~Chang, J.~Lu, F.~Yu, and A.~Finkelstein.
\newblock Pairedcyclegan: Asymmetric style transfer for applying and removing
  makeup.
\newblock In {\em IEEE Conference on Computer Vision and Pattern Recognition},
  2018.

\bibitem{chen2016ensemble}
C.~Chen, A.~Dantcheva, and A.~Ross.
\newblock An ensemble of patch-based subspaces for makeup-robust face
  recognition.
\newblock {\em Information Fusion}, 32:80--92, 2016.

\bibitem{dantcheva2012can}
A.~Dantcheva, C.~Chen, and A.~Ross.
\newblock Can facial cosmetics affect the matching accuracy of face recognition
  systems?
\newblock In {\em the Fifth International Conference on Biometrics: Theory,
  Applications and Systems}, pages 391--398. IEEE, 2012.

\bibitem{goodfellow2014generative}
I.~Goodfellow, J.~Pouget-Abadie, M.~Mirza, B.~Xu, D.~Warde-Farley, S.~Ozair,
  A.~Courville, and Y.~Bengio.
\newblock Generative adversarial nets.
\newblock In {\em Advances in neural information processing systems}, pages
  2672--2680, 2014.

\bibitem{guo2014face}
G.~Guo, L.~Wen, and S.~Yan.
\newblock Face authentication with makeup changes.
\newblock {\em IEEE Transactions on Circuits and Systems for Video Technology},
  24(5):814--825, 2014.

\bibitem{guo2016ms}
Y.~Guo, L.~Zhang, Y.~Hu, X.~He, and J.~Gao.
\newblock Ms-celeb-1m: A dataset and benchmark for large-scale face
  recognition.
\newblock In {\em European Conference on Computer Vision}, pages 87--102.
  Springer, 2016.

\bibitem{he2016deep}
K.~He, X.~Zhang, S.~Ren, and J.~Sun.
\newblock Deep residual learning for image recognition.
\newblock In {\em the IEEE conference on computer vision and pattern
  recognition}, pages 770--778, 2016.

\bibitem{he2017learning}
R.~He, X.~Wu, Z.~Sun, and T.~Tan.
\newblock Learning invariant deep representation for nir-vis face recognition.
\newblock In {\em The Thirty-First AAAI Conference on Artificial Intelligence},
  pages 2000--2006. AAAI Press, 2017.

\bibitem{hu2013makeup}
J.~Hu, Y.~Ge, J.~Lu, and X.~Feng.
\newblock Makeup-robust face verification.
\newblock In {\em International Conference on Acoustics, Speech and Signal
  Processing}, pages 2342--2346, 2013.

\bibitem{isola2016image}
P.~Isola, J.-Y. Zhu, T.~Zhou, and A.~A. Efros.
\newblock Image-to-image translation with conditional adversarial networks.
\newblock In {\em IEEE Conference on Computer Vision and Pattern Recognition},
  pages 5967--5976. IEEE, 2017.

\bibitem{jing2016multi}
X.-Y. Jing, F.~Wu, X.~Zhu, X.~Dong, F.~Ma, and Z.~Li.
\newblock Multi-spectral low-rank structured dictionary learning for face
  recognition.
\newblock {\em Pattern Recognition}, 59:14--25, 2016.

\bibitem{johnson2016perceptual}
J.~Johnson, A.~Alahi, and L.~Fei-Fei.
\newblock Perceptual losses for real-time style transfer and super-resolution.
\newblock In {\em European Conference on Computer Vision}, pages 694--711.
  Springer, 2016.

\bibitem{li2018global}
P.~Li, Y.~Hu, Q.~Li, R.~He, and Z.~Sun.
\newblock Global and local consistent age generative adversarial networks.
\newblock {\em arXiv preprint arXiv:1801.08390}, 2018.

\bibitem{li2018beautygan}
T.~Li, R.~Qian, C.~Dong, S.~Liu, Q.~Yan, W.~Zhu, and L.~Lin.
\newblock Beautygan: Instance-level facial makeup transfer with deep generative
  adversarial network.
\newblock In {\em 2018 ACM Multimedia Conference on Multimedia Conference},
  pages 645--653. ACM, 2018.

\bibitem{li2017anti}
Y.~Li, L.~Song, X.~Wu, R.~He, and T.~Tan.
\newblock Anti-makeup: Learning a bi-level adversarial network for
  makeup-invariant face verification.
\newblock In {\em The Thirty-Second AAAI Conference on Artificial
  Intelligence}, 2018.

\bibitem{li2019learning}
Y.~Li, L.~Song, X.~Wu, R.~He, and T.~Tan.
\newblock Learning a bi-level adversarial network with global and local
  perception for makeup-invariant face verification.
\newblock {\em Pattern Recognition}, 2019.

\bibitem{Mejjati2018Unsupervised}
Y.~A. Mejjati, C.~Richardt, J.~Tompkin, D.~Cosker, and K.~I. Kim.
\newblock Unsupervised attention-guided image to image translation.
\newblock In {\em Thirty-second Conference on Neural Information Processing
  Systems}, 2018.

\bibitem{nguyen2010cosine}
H.~V. Nguyen and L.~Bai.
\newblock Cosine similarity metric learning for face verification.
\newblock In {\em Asian Conference on Computer Vision}, pages 709--720.
  Springer, 2010.

\bibitem{ronneberger2015u}
O.~Ronneberger, P.~Fischer, and T.~Brox.
\newblock U-net: Convolutional networks for biomedical image segmentation.
\newblock In {\em International Conference on Medical Image Computing and
  Computer-Assisted Intervention}, pages 234--241. Springer, 2015.

\bibitem{ruprecht1995image}
D.~Ruprecht and H.~Muller.
\newblock Image warping with scattered data interpolation.
\newblock {\em IEEE Computer Graphics and Applications}, 15(2):37--43, 1995.

\bibitem{simonyan2014very}
K.~Simonyan and A.~Zisserman.
\newblock Very deep convolutional networks for large-scale image recognition.
\newblock In {\em Proceedings of the 3rd International Conference on Learning
  Representations}, 2015.

\bibitem{sun2014deep}
Y.~Sun, Y.~Chen, X.~Wang, and X.~Tang.
\newblock Deep learning face representation by joint
  identification-verification.
\newblock In {\em Advances in neural information processing systems}, pages
  1988--1996, 2014.

\bibitem{sun2017weakly}
Y.~Sun, L.~Ren, Z.~Wei, B.~Liu, Y.~Zhai, and S.~Liu.
\newblock A weakly supervised method for makeup-invariant face verification.
\newblock {\em Pattern Recognition}, 66:153--159, 2017.

\bibitem{sun2013hybrid}
Y.~Sun, X.~Wang, and X.~Tang.
\newblock Hybrid deep learning for face verification.
\newblock In {\em the IEEE International Conference on Computer Vision}, pages
  1489--1496, 2013.

\bibitem{taigman2014deepface}
Y.~Taigman, M.~Yang, M.~Ranzato, and L.~Wolf.
\newblock Deepface: Closing the gap to human-level performance in face
  verification.
\newblock In {\em the IEEE Conference on Computer Vision and Pattern
  Recognition}, pages 1701--1708, 2014.

\bibitem{wei2017learning}
Z.~Wei, Y.~Sun, J.~Wang, H.~Lai, and S.~Liu.
\newblock Learning adaptive receptive fields for deep image parsing network.
\newblock In {\em Proceedings of the IEEE Conference on Computer Vision and
  Pattern Recognition}, pages 2434--2442, 2017.

\bibitem{wu2018light}
X.~Wu, R.~He, Z.~Sun, and T.~Tan.
\newblock A light cnn for deep face representation with noisy labels.
\newblock {\em IEEE Transactions on Information Forensics and Security},
  13(11):2884--2896, 2018.

\bibitem{zhang2016multi}
S.~Zhang, R.~He, Z.~Sun, and T.~Tan.
\newblock Multi-task convnet for blind face inpainting with application to face
  verification.
\newblock In {\em International Conference on Biometrics}, pages 1--8, 2016.

\bibitem{Zhu2017Unpaired}
J.-Y. Zhu, T.~Park, P.~Isola, and A.~A. Efros.
\newblock Unpaired image-to-image translation using cycle-consistent
  adversarial networks.
\newblock In {\em IEEE International Conference on Computer Vision}, 2017.

\end{thebibliography}
}

\end{document}